# Multi-task Collaborative Pre-training and Individual-adaptive-tokens Fine-tuning: A Unified Framework for Brain Representation Learning

Ning Jiang, Gongshu Wang, and Tianyi Yan

*Abstract*—**Structural magnetic resonance imaging (sMRI) provides accurate estimates of the brain's structural organization and learning invariant brain representations from sMRI is an enduring issue in neuroscience. Previous deep representation learning models ignore the fact that the brain, as the core of human cognitive activity, is distinct from other organs whose primary attribute is anatomy. Therefore, capturing the semantic structure that dominates interindividual cognitive variability is key to accurately representing the brain. Given that this high-level semantic information is subtle, distributed, and interdependently latent in the brain structure, sMRI-based models need to capture fine-grained details and understand how they relate to the overall global structure. However, existing models are optimized by simple objectives, making features collapse into homogeneity and worsening simultaneous representation of fine-grained information and holistic semantics, causing a lack of biological plausibility and interpretation of cognition. Here, we propose MCIAT, a unified framework that combines Multi-task Collaborative pre-training and Individual-Adaptive-Tokens fine-tuning. Specifically, we first synthesize restorative learning, age prediction auxiliary learning and adversarial learning as a joint proxy task for deep semantic representation learning. Then, a mutual-attention-based token selection method is proposed to highlight discriminative features. The proposed MCIAT achieves state-of-the-art diagnosis performance on the ADHD-200 dataset compared with several sMRI-based approaches and shows superior generalization on the MCIC and OASIS datasets. Moreover, we studied 12 behavioral tasks and found significant associations between cognitive functions and MCIAT-established representations, which verifies the interpretability of our proposed framework.**

## I. INTRODUCTION

LEARNING general and robust brain representations is a critical and enduring problem in neuroimaging that underlies many neuroscientific analyses (i.e., brain disease diagnosis, structural segmentation, brain pattern decoding, brain structure-behavior association establishment, etc.). Previous studies have focused on manually extracting morphological estimates of brain structure, such as grey matter (GM) volume, cortical thickness, and surface area [1-3]. To obtain more meaningful representations, deep brain representation learning has recently gained traction and has been widely used in brain imaging analysis [4-6]. These approaches tend to follow a similar research paradigm: combining a commonly used deep representation learning model (generative adversarial networks (GANs) [7], variational autoencoder (VAE) [8], etc.) with a biologically meaningful prior, e.g., topological invariance of the anatomical structure [4], which is inserted into deep learning (DL) frameworks as a customized regularization or module. Afterward, they verify the generalization of learned representations via multiple approaches, including downstream tasks (e.g., Alzheimer's disease (AD) diagnosis, brain tumor segmentation, modality synthesis), location of important brain regions, visualization of feature maps, and inter-group comparison, which resonate with the analysis in other medical data domains [9].

However, the brain is made up of countless interacting neurons, which is fundamentally different from other organs. The structural basis of the brain dominates a wide range of organismic behaviors. The structural alterations, represented by "visual semantic changes" in images of organs such as the lungs and heart, are of anatomical significance only. However, in the brain, minor structural alterations may cause significant cognitive changes or abnormalities. In clinical practice, such changes or abnormalities often occur long before but are not identified until obvious cognitive deficits have developed, and in some cases remain undetected for a lifetime. Therefore, capturing representations that can encode rich cognitive information and reflect the brain's built-in patterns is key to brain representation learning. This can guide not only cognitive neuroscience research but also clinical application.

Considering that the complex mechanisms of human cognition rely on distributed neural networks and are generally covariant, the structural variations that dominate interindividual cognitive variations are naturally subtle, mixed, and inter-dependent. To fully represent these informative and distinctive features, DL models need to capture fine-grained details and understand how they relate to the overall global structure. In other words, the fine-grained recognition capability on one hand, and the holistic semantic understanding of input images on the other, should be considered simultaneously when designing a deep brain representation model.

Therefore, the existing studies have the following two limitations: (1) The simple training tasks are not well adapted to the biological characteristics of the brain structure, which is not conducive to capturing both detailed and global semantic information. (2) The explained variance of brain representations for cognition is neglected. Their results of normal representation analysis, such as localization to cognitively relevant brain regions or inter-group differences between patients and healthy people, are superficial and nonindividual and thus not sufficient to demonstrate the biological feasibility of the learned representations.

A series of studies have confirmed the power of the Vison Transformer (ViT) [10] to capture both global and local features. This ability makes ViT innately suitable for representation learning of brain structure. The long-range "receptive field" of ViT enables it to locate subtle changes and establish their dependencies, thus highlighting fine-grained features to address the abovementioned challenges. We propose a ViT-based MCIAT, which leverages restorative learning, age prediction auxiliary learning, and adversarial learning to foster Multi-task Collaborative pre-training and combines Individual-Adaptive-Tokens fine-tuning realized by a mutual attention mechanism.

Specifically, we first reshape an input structural magnetic resonance imaging (sMRI) into several patches, and a random mask divides the patches into visible patches and masked patches. The pre-training stage has four proxy tasks: (1) Infer the mask target from the visible context, forcing the model to build long-range dependencies and understand contextual semantic information. (2) Restore the distorted visible patches back to the original pixels, encouraging the model to focus on the structural information hidden in image details. (3) Predict the age of the input sample from the visible patches, naturally revealing the biological information generated during brain development and aging. (4) Reinforce the restoration quality by adversarial learning, making it essential for the model to establish a latent space that matches the distribution of brain sMRI to fool the discriminator with images reconstructed from the latent space. Through the synergy and co-training of the four proxy tasks, the model is incorporated into an effective constraint and is optimized for a more complex and difficult objective, which facilitates simultaneous learning of detailed and global semantic information. A shared ViT encoder, three independent lightweight decoders, and a convolutional neural network (CNN) discriminator are used for pre-training. Then, we transfer the pre-trained encoder to downstream tasks. For fine-tuning, we propose a mutual-attention-based method to adaptively and individually highlight and select important tokens from each transformer layer. Finally, the selected tokens are fed into the classifier as identified discriminative features. Three challenging downstream tasks are leveraged to validate the capacity of our proposed MCIAT: (1) Attention deficit hyperactivity disorder (ADHD) diagnosis. (2) Schizophrenia (SZ) diagnosis. (3) Preclinical AD identification. Unlike neurological diseases, e.g., AD, with obvious morphological changes caused by brain atrophy, psychiatric diseases, such as ADHD and SZ, show subtle structural alterations and thus are difficult to distinguish. For subjects we defined as "preclinical AD", all the parameters, physical assessments, and clinical data state that the patient is healthy and showing no symptoms, but brain neuronal structure has started to deteriorate [11]. The MCIAT outperforms several compared methods in challenging downstream tasks, showing superiority in representation learning and fine-grained categorization. Furthermore, we verify the interpretability of the proposed model using various methods, including not only common representation analysis techniques but also verifying the associations between brain representations and cognitive functions and analyzing the interindividual variability of learned representations. Our main contributions are summarized as follows:

(1) We propose MCIAT, a unified framework that combines multi-task collaborative pre-training and individual-adaptive-tokens fine-tuning, which exhibits superior performance in downstream tasks.

(2) We studied 12 behavioral tasks to explore the associations between deep brain representation and behavioral functions. We used multiple representation analysis and visualization methods to verify the interpretability of our MCIAT and demonstrate that the proposed model represents the semantic information relevant to cognitive functions involving comprehension, reasoning, memory, emotion, and learning.

(3) We explore how different types of pre-trained components affect the representation learning capacity of ViT, revealing the promising potential of ViT for both competitive performance and interpretability. Such endeavors, we envision, will inspire future studies of tailor-made deep brain representation learning frameworks using ViT as the backbone instead of CNNs.

## II. RELATED WORK

### A. Multi-task Learning of Brain Representations

Multi-task learning (MKL) and collaborative learning have been widely employed in visual semantic learning [12-14], and there have been attempts to deploy MKL for medical image analysis, especially for medical segmentation [15, 16]. In recent years, MKL has also been involved in brain representation learning, such as brain stroke lesion segmentation [17], brain tumor segmentation [18, 19], and brain disease diagnosis [20]. Specifically, [17] adopted stroke lesion mask-prediction as the main task and stroke lesion edge-prediction as the auxiliary task for segmentation of stroke lesions. [18] integrated three tasks (i.e., coarse segmentation of the complete tumor, refined segmentation for the complete tumor and its intra-tumoral classes and precise segmentation for the enhancing tumor) into one framework and achieved state-of-the-art performance on two datasets. [19] utilized a shared encoder to extract features and two task-specific decoders, a segmentation decoder and an auxiliary fusion decoder for better segmentation performance. [20] added two tasks (prediction of the Mini-Mental State Examination (MMSE) and Alzheimer's Disease Assessment Scale-Cognitive Subscale (ADAS-Cog)) to their basic AD diagnosis task for higher classification accuracy.

The abovementioned multi-task brain representation learning approaches, in fact, trained their models via multiple tasks but were limited in a single task domain. For example, all the tasks in [17] and [18] are mask-prediction tasks whose essence is pixel-classification, and all the tasks in [20] are label-prediction tasks. given that tasks of different domains (e.g., restorative learning, adversarial learning) have different implications for representation learning, this limitation may hinder their models from learning effective and diverse brain representations.

### B. Attention-Guided Feature Selection

Inspired by the successes of attention mechanisms in CNNs, several studies have employed attention guidance to identify anatomically meaningful regions in the brain [21-23]. The H-FCN [21] generated location proposals using anatomical biomarkers as prior knowledge and then selected discriminative regions from a whole brain MRI via an attention mechanism for accurate AD diagnosis. Similar to [21], HybNet [22] also selected discriminative regions for AD diagnosis, but its feature selection was guided by a class activation map (CAM), which was extracted from pre-trained models. In addition, the selected multi-scale features were further fused for the classification task. The DA-MIDL [23] model consists of patch-net with a spatial attention block, attention multi-instance learning pooling and an attention-aware global classifier to construct an effective classification model for AD diagnosis.

However, existing attention-guided methods were designed for CNNs and introduced extra parameters thus are inapplicable to ViT based models. At the core of ViT, the self-attention mechanism may be inherently capable to generate guidance for feature selection, without introducing any extra parameters.

### C. Linking Brain Structure to Behavior

Understanding the brain structural correlations of inter-individual differences in behavior is an important objective in cognitive neuroscience. Recently, searching for associations between local measurements of neuroanatomy and behavioral variables has gained traction in brain structure-behavior studies [24-26]. [24] attempted to link the Big Five personality traits of openness-to-experience with variability in brain structural features (cortical thickness, surface area, subcortical volume and white matter microstructural integrity) while producing inconsistent results. [25] studied how children's family income gaps in cognition relate to the volume of the anterior and posterior hippocampus. They observed that the anterior, but not posterior hippocampus mediates income-related differences in cognitive scores. [26] used a fully probabilistic approach to explore the variation in regional GM volume depending on sex and social traits and provide a population-level window into the brain associations with social behavior. These methods usually use manual features, whereas the high-level features extracted by DL models are coarse, abstract, and hard to understand at the level of human knowledge; hence, current research on the link between deep brain representation and behavior is scarce.

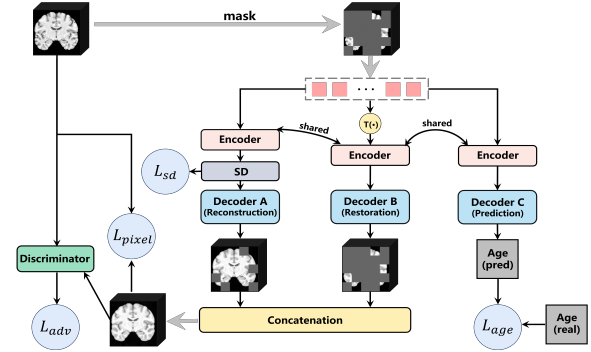

Fig. 1. Pre-training framework, where SD denotes semantic diversity regularization and $T(\cdot)$ denotes a transformation function.

## III. METHOD

### A. MCIAT Framework

#### 1) Collaborative Pre-training

**Restorative Learning.** Our restorative learning framework consists of two branches: parallel distillation masked image modeling (named branch A) and visible portion restoration (named branch B). Parallel distillation, using semantic diversity regularization to guide masked image modeling, was shown to be effective to learn the brain's built-in pattern in our previous work [27]. Therefore, we introduce branch A to encourage the encoder to distill deep semantic information for comprehensive understanding of the brain. Conversely, branch B is exploited to enhance fine-grained representation learning, especially for subtle, local variations. As shown in Fig. 1, we first reshape an input sMRI $x$ into 3D patches $x_{all}$ without overlapping. Then, random masking divides $x_{all}$ into two portions: $x_{vis}$ and $x_{mask}$.

For branch A, $x_{vis}$ is processed by the shared encoder $E$ to generate the latent feature $y_{vis} = E(x_{vis})$. The projection heads $h_{A1}$ and $h_{A2}$ map $y_{vis}$ into latent embeddings $z_{vis\_A1}$ and $z_{vis\_A2}$. We expect the two latent embeddings to represent opposing semantic information, and a semantic diversity loss is designed to disentangle the two in the latent space. Specifically, we compute the Gram matrices $G_{vis\_A1} = z_{vis\_A1} \cdot z^T_{vis\_A1}$ and $G_{vis\_A2} = z_{vis\_A2} \cdot z^T_{vis\_A2}$, and the semantic diversity loss is:

$$L_{sd} = -log(\sigma(G_{vis_{A1}})) - log(1 - \sigma(G_{vis_{A2}})) \quad (1)$$

All of the tokens are fed into the reconstruction decoder $D_A$, including $z_{vis\_A1}$, $z_{vis\_A2}$, and the masked tokens $z_{mask}$, which are learnable tokens put in the positions of $x_{mask}$, and the reconstructed masked patches are output:

$$x_{mask\_recon} = D_A(z_{vis_{A1}} \oplus z_{mask}) - D_A(z_{vis_{A2}} \oplus z_{mask}) \quad (2)$$

where $\oplus$ denotes the concatenation operation of tensors.

For branch B, we first distort the visible patches $x_{vis}$ by a function $T$. The goal of the encoder $E$, the projection head $h_B$ and the restoration decoder $D_B$ is to map the distorted patches back to the original ones. The output can be described as:

$$x_{vis\_restore} = D_B(h_B(E(T(x_{vis})))) \quad (3)$$

Note that $T$ is a nonlinear transformation function realized by Bézier curves [28].

Finally, an entire image is assembled with the reconstructed masked patches and the restored visible patches:

$$x_{full\_restore} = x_{mask\_recon} + x_{vis\_restore} \quad (4)$$

The restorative loss can be described as:

$$L_{pixel} = MSE(x_{full\_restore}, x) \quad (5)$$

**Age Prediction Auxiliary Learning.** Age information is easily available, and biologically meaningful phenotypes emerge naturally through age prediction [29]. Semantics associated with aging and development are also related to cognition and pathology [30]. The goal of $E$ and the age-prediction decoder $D_{age}$ is to predict the age of the input sample using the visible patches $x_{vis}$, which can be described as:

$$x_{age\_pred} = D_{age}(E(x_{vis})) \quad (6)$$

$$L_{age} = MSE(x_{age\_pred}, x_{age\_target}) \quad (7)$$

**Adversarial Learning.** Adversarial learning exploits a discriminator $D_{adv}$ to distinguish the restored images from the original ones. Benefiting from the minimax game between $E$ and $D_{adv}$, the encoder can better represent subtle structural changes, which are informative and discriminative for capturing cognitive variations. The adversarial loss [7] is:

$$L_{adv} = \log(D_{adv}(x)) + \log(1 - D_{adv}(x_{full\_restore})) \quad (8)$$

**Collaborative Pre-training.** Finally, the overall objective of pre-training becomes:

$$L = \lambda_{sd} * L_{sd} + \lambda_{pixel} * L_{pixel} + \lambda_{age} * L_{age} + \lambda_{adv} * L_{adv} \quad (9)$$

where $\lambda_{sd}$, $\lambda_{pixel}$, $\lambda_{age}$ and $\lambda_{adv}$ are hyperparameters that determine the importance of losses. Through collaborative pre-training, $E$ can represent fine-grained semantics and establish the dependency between fine-grained details and the global context. In particular, $\lambda_{sd}$ helps the model more deeply learn diverse semantic representations. $\lambda_{pixel}$ forces the model to infer the masked target by aggregating visible context and restoring the distorted pixels in visible patches, which is beneficial for understanding the global context and capturing fine-grained features. $\lambda_{adv}$ reinforces restorative learning to capture more significant features. Finally, $\lambda_{age}$ enhances the biological plausibility of the learned representations.

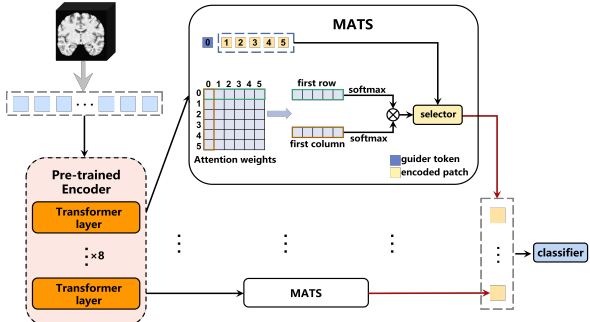

Fig. 2. Fine-tuning framework, where MATS denotes the mutual-attention-based token selection module.

### 2) Individual-Adaptive-Tokens Fine-Tuning

For fine-tuning, a new classifier with initialized weights is added to the pre-trained encoder, which can extract subtle yet discriminative features but needs to further highlight and strengthen task-specific features in fine-tuning. We propose a mutual-attention-based token selection (MATS) method and insert it into the pre-trained ViT backbone, where we adaptively and individually select the informative tokens from each transformer layer (see Fig. 2 for more details).

Our proposed token selection method introduces an additional independent guider token $G$ to identify important tokens. Similar to the class token in [10], the initialized $G$ does not contain any information about the input but interacts with all patch tokens layer by layer. Unlike ViT, $G$ is only used to guide token-selection and is not fed into the classifier. The input of the classifier is the set of selected tokens. We are motivated by this assumption: the guider token $G$ aggregates the global context from all patch tokens, thus containing information that can describe the holistic semantics and style of the input sMRI. Meanwhile, $G$ stays out of the classifier, so it has no preference for tokens containing more low-level structural information but becomes an impartial guider that is rich in global information. Therefore, a set of tokens, selected from each transformer layer, which is the most relative to $G$, can be seen as the most discriminative representation of the input. A mutual attention mechanism is used to compute the importance score of tokens:

$$A = \begin{bmatrix} a_{00} & \cdots & a_{0j} & \cdots & a_{0N} \\ \vdots & \ddots & \vdots & \ddots & \vdots \\ a_{i0} & \cdots & a_{ij} & \cdots & a_{iN} \\ \vdots & \ddots & \vdots & \ddots & \vdots \\ a_{N0} & \cdots & a_{Nj} & \cdots & a_{NN} \end{bmatrix} \quad (10)$$

$$r_i = [a_{i0}, a_{i1}, a_{i2}, \ldots, a_{ij}, \ldots, a_{iN}] \quad (11)$$

$$c_j = [a_{0j}, a_{1j}, a_{2j}, \ldots, a_{i,j}, \ldots, a_{Nj}]^T \quad (12)$$

where $A \in \mathbb{R}^{(N+1)*(N+1)}$ denotes an attention weight matrix for one attention head, and $N$ is the number of tokens. $r_i$ denotes row $i$ of $A$, $c_j$ denotes column $j$ of $A$, and $a_{i,j}$ is the attention weight between token $i$ and token $j$ in the context of token $i$. We define the guider token $G$ as token $0$, and the mutual attention score $S_i$ between token $G$ and token $i$ is:

$$S_i = \frac{e^{a_{0i}}}{\sum_{j=0}^{N} e^{a_{0j}}} * \frac{e^{a_{i0}}}{\sum_{j=0}^{N} e^{a_{j0}}} \quad (13)$$

We average the attention weights of all the heads to compute the mutual attention scores. Then, the tokens are sorted by their scores, and we select the top tokens from each layer according to their indices. Finally, all selected tokens from a sample $x$ are fed into the classifier, but not the guider token $G$.

### A. Experiments

### 1) Dataset Description and Data Partition

The brain structural images (T1w) were provided by the Cambridge Centre for Aging and Neuroscience (Cam-CAN) [31], ADHD-200 [32], MIND Clinical Imaging Consortium

(MCIC) [33] and OASIS [34]. We performed image preprocessing using sMRIprep. The T1w image was corrected for intensity nonuniformity with N4 bias correction and was skull-stripped using the antsBrainExtraction workflow. Volume-based spatial normalization to standard space (MNI152NLin2009cAsym, 181×217×181 voxels) was performed through nonlinear registration with antsRegistration. Then, we cut off part of the black background at the edges, and a main area (150×180×150 voxels) was reserved.

In this study, 636 cognitively healthy adults from Cam-CAN were used for collaborative pre-training. For ADHD-200, we used the same general training/testing sets as in previous studies: 768 training subjects (488 normal control subjects (NCs), 280 ADHD patients) and 171 test subjects (94 NCs, 77 ADHD patients) collected from eight independent sites. For the MCIC and OASIS datasets, we performed ten-fold cross-validation on 190 subjects (99 NCs, 91 SZ patients) and 280 subjects (190 NCs, 90 preclinical ADs) and reported the averaged accuracy (ACC), sensitivity (SEN), specificity (SPE), and area under the receiver operating characteristic (ROC) curve (AUC).

### 2) Experimental Settings

**Training Protocol.** In pre-training, the model is optimized by AdamW with a batch size of 16 and a weight decay parameter of 0.05 on four NVIDIA Tesla P100 GPUs using PyTorch. The model is pre-trained for 3000 epochs (40 warm-up epochs) with a cosine linear-rate scheduler, and the initial learning rate is set as $lr \times BatchSize/4$, where $lr = 1.5e^{-4}$. We empirically set the mask ratio to 0.76, the patch size to 30×30×30, the nonlinear rate for the function $T$ to 0.9, and $\lambda_{sd}, \lambda_{pixel}, \lambda_{age}, \lambda_{adv}$ to 0.005, 0.79, 0.1, and 0.1, respectively. The backbone of the shared encoder $E$ is a ViT composed of 8 transformer blocks using 12 heads for multi-head self-attention (MSA) and 512 as the hidden size. The decoders $D_A$, $D_B$ and $D_{age}$ are composed of a single transformer block using 12 heads for MSA and 512 as the hidden size. The discriminator $D_{adv}$ is ResNet10, which is strictly the same as in [35].

In fine-tuning, we transfer the pre-trained encoder $E$ to downstream tasks by fine-tuning all of the parameters. Unlike previous methods [36-38], each layer is equally important in MCIAT, so layer-decay is not used in fine-tuning.

The search space of pre-training hyperparameters, especially for $\lambda_{sd}, \lambda_{pixel}, \lambda_{age}$ and $\lambda_{adv}$, and fine-tuning hyperparameters for each downstream task are listed in our codebase. [1]

**Baselines.** Our proposed MCIAT is compared with four methods, including two general baselines (ResNet50 [35] and ViT [10]) and two state-of-the-art models designed for a specific brain disease (DA-MIDL [23] and Incep_Resnet [39]).

## IV. RESULTS AND DISCUSSION

### A. Classification Performance

The classification results of ADHD-200 are shown in Table I. The MCIAT achieves the best accuracy (74.27%) compared to baselines that adopted the official partition of training/testing sets using sMRI, with up to 5.26% improvement in accuracy. Our proposed MCIAT even surpasses most of the approaches

[1] Codebase: https://github.com/NingJiang-git/MCIAT

### TABLE I
EXPERIMENTAL RESULTS (%) ON ADHD-200

| Method | Features | ACC | SEN | SPE | AUC |
|---|---|---|---|---|---|
| Quotes results from the literature (sMRI-based) | | | | | |
| *3D-CNN[40]* | *GM density* | *65.86* | *-* | *-* | *-* |
| *Tensor Boosting[41]* | *sMRI* | *69.01* | *-* | *-* | *-* |
| *3D-CNN[2]* | *FDCM of GM density* | *69.01* | *-* | *-* | *-* |
| Quotes results from the literature (fMRI-based) | | | | | |
| *3D-CNN[40]* | *fALFF* | *66.04* | *-* | *-* | *-* |
| *FC-HAT[42]* | *FBN* | *69.2* | *83.0* | *46.8* | *-* |
| *EM-MI[43]* | *fMRI* | *70.4* | *-* | *-* | *-* |
| *4D-CNN[44]* | *fMRI* | *71.3* | *-* | *-* | *-* |
| *SASNI[45]* | *fMRI* | *72.5* | *-* | *-* | *-* |
| Quotes results from the literature (Multimodality) | | | | | |
| *MKL[46]* | *ReHo, GM and cortical thickness* | *61.54* | *-* | *-* | *-* |
| *3D-CNN[40]* | *fALFF and GM density* | *69.15* | *-* | *-* | *-* |
| *MGF[47]* | *sMRI, fMRI* | *71.9* | *-* | *-* | *-* |
| Results obtained by this paper | | | | | |
| ReNet50 | sMRI | 69.01 | 57.14 | 78.72 | 65.06 |
| ViT | sMRI | 64.91 | 40.26 | 85.11 | 64.49 |
| DA-MIDL | sMRI | 61.99 | **62.34** | 61.70 | 62.59 |
| Incep_Resnet | sMRI | 65.50 | 38.96 | **87.23** | 69.23 |
| MCIAT$_1$ | sMRI | 71.35 | 53.25 | 86.17 | 74.79 |
| MCIAT$_2$ | sMRI | 72.51 | 55.84 | 86.17 | 74.14 |
| MCIAT$_3$ | sMRI | **74.27** | 61.04 | 85.11 | 72.07 |
| MCIAT$_4$ | sMRI | 73.68 | 57.14 | **87.23** | **75.79** |

\* The rows in italics have their numbers copied from the studies using the same training/testing sets as ours. Note that MCIAT$_N$ denotes that the model adaptively selects $N$ tokens per layer in fine-tuning.

from the literature that use functional features of the brain (usually functional connectivity estimates) or multimodality features, such as 4D-CNN, SASNI and MGF.

The effectiveness of the number of selected tokens is shown in Table I. For ADHD classification, our method achieves the best performance when selecting three tokens per layer, which is the default setting in subsequent experiments. Note that MCIAT always reports a competitive accuracy (over 71.35%) with different settings, indicating that the model has established general and robust representations for brain sMRI images.

### TABLE II
EXPERIMENTAL RESULTS (%) ON MCIC

| Method | Features | ACC | SEN | SPE | AUC |
|---|---|---|---|---|---|
| ReNet50 | sMRI | 77.37±7.87 | 78.00±22.50 | 76.67±19.21 | 69.22±16.14 |
| ViT | sMRI | 70.00±9.62 | **81.00±11.01** | 57.78±20.15 | 72.89±9.27 |
| DA-MIDL | sMRI | 64.74±6.10 | 64.00±22.71 | 65.56±17.72 | 66.25±10.29 |
| Incep_Resnet | sMRI | 72.63±6.93 | 73.00±18.89 | 72.22±13.10 | 69.44±10.05 |
| MCIAT$_3$ | sMRI | **80.00±6.47** | 79.00±13.70 | **81.11±12.88** | **78.67±11.28** |

### TABLE III
EXPERIMENTAL RESULTS (%) ON OASIS

| Method | Features | ACC | SEN | SPE | AUC |
|---|---|---|---|---|---|
| ReNet50 | sMRI | 78.93±6.62 | 62.22±23.54 | 76.67±10.89 | 74.21±17.75 |
| ViT | sMRI | 74.64±2.63 | 45.68±16.94 | 87.72±8.52 | 63.71±12.34 |
| DA-MIDL | sMRI | 75.00±5.32 | 50.00±24.71 | 86.84±7.94 | 61.08±15.57 |
| Incep_Resnet | sMRI | 78.21±8.82 | 55.56±19.60 | 86.84±7.55 | 77.95±9.62 |
| MCIAT$_3$ | sMRI | **83.57±1.84** | **71.61±19.03** | **89.47±7.85** | **79.98±8.05** |

The classification results of the MCIC and OASIS datasets are shown in Table II and Table III. The impressive performance demonstrates the generalization of our proposed MCIAT and suggests that the model establishes a strong latent space that semantically interprets the brain's built-in pattern to encode those implicit features into an explicit form. As mentioned in Section I, the capacity of MCIAT to discriminate preclinical AD from healthy people demonstrates that MCIAT has captured the deterioration of brain neuronal structure at its nascent stage, which has important clinical significance.

TABLE IV
ABLATION STUDY RESULTS (%) ON DIFFERENT COMPONENTS OF MCIAT

| Ablation Mode | Pre-training. | | | | Fine-tuning. | ACC | | |
|---|---|---|---|---|---|---|---|---|
| | $Res_A$ | Age | $Res_B$ | Adv | IAT | ADHD-200 | MCIC | OASIS |
| mode0 | × | × | × | × | ✓ | 65.50 | 74.21±4.61 | 74.64±2.63 |
| mode1 | ✓ | × | × | × | ✓ | 66.08↑ | 77.37±7.47↑ | 77.86±3.69↑ |
| mode2 | ✓ | ✓ | × | × | ✓ | 66.67↑ | 78.42±8.02↑ | 79.64±2.94↑ |
| mode3 | ✓ | ✓ | ✓ | × | ✓ | 70.18↑ | 76.32±7.55↓ | 79.28±4.70↓ |
| mode4 (w/o IAT) | ✓ | ✓ | ✓ | ✓ | × | 69.01 | 78.42±8.02 | 78.21±4.59 |
| mode4 (proposed) | ✓ | ✓ | ✓ | ✓ | ✓ | **74.27↑** | **80.00±6.47↑** | **83.57±1.84↑** |

* $Res_A$ and $Res_B$ denote restorative learning branches A and B. IAT denotes individual adaptive tokens in fine-tuning.

### B. Ablation Study

We perform a thorough ablation study to show how each component contributes to MCIAT. For pre-training, we start with a ViT backbone and incrementally add restorative learning branch A, age prediction auxiliary learning, restorative learning branch B, and adversarial learning. For fine-tuning, we compare MCIAT with ViT without individual adaptive tokens when all four pre-training components are applied.

The ablation results are shown in Table IV. We make the following observations: (1) In pre-training, each component significantly improves the performance of ADHD diagnosis, while in SZ diagnosis and preclinical AD identification, there is slight performance degradation when adding restorative learning branch B. Note that this gap is later compensated after adding adversarial learning, which indicates that collaborative pre-training, especially the unification of restorative learning and adversarial learning, achieves a complementary harmony and thus has superior performance in downstream tasks. This conclusion could be verified by ablation on ADHD-200. After adding adversarial learning, there is an obvious improvement in accuracy (+4.09%). (2) The individual-adaptive-tokens strategy enhances all of the pre-trained models in downstream tasks. The proposed MCIAT is an end-to-end model with T1w images as its input and automatically and adaptively selects features across individuals. The ablation results demonstrate the effectiveness of our proposed MATS method.

To provide theoretical guidance for subsequent research on how to tailor a brain representation learning framework for ViT to specific task/data domains, we further explore the following three issues: (1) Are the representations extracted by the pre-trained models associated with cognitive functions? How can the biological interpretability of the model be verified? (2) For each of the components in pre-training, what representations are promoted, and what representations are damaged? (3) For medical image analysis, is feature selection and fusion for different transformer layers (i.e., individual-adaptive-tokens strategy) a more appropriate approach than using features from the last layer (i.e., vanilla ViT) for downstream tasks?

### C. Representation Analysis for the Pre-trained models
#### 1) Association between Representation and Cognition

Partial least square regression (PLSR) is widely used to establish brain-behavior relationships [30]. In this study, we employed PLSR to build prediction models for individual brain representations and behavior metric scores. We study 12 behavioral tasks (Emotion Memory, Fluid Intelligence (FI), Picture Priming, Motor Learning, Sentence Comprehension, Face Recognition: Familiar Faces, Face Recognition: Unfamiliar Faces, Emotional Regulation, Visual Short-Term Memory (VSTM), RT Simple, Force Matching and Tip-of-the-Tongue Task (TOT)), which are described in detail by [48] (see the study protocol in [49]), and follow the definition in [50] that RT Simple, Force Matching and TOT are basic active tasks and the other nine are cognitive tasks.

Specifically, for all subjects in the Cam-CAN dataset, we first freeze the parameters of our pre-trained encoder and three pre-trained ablation models and extract the encoded features from each transformer layer during a forward propagation process. Then, the encoded features are used to predict each metric score of the 12 behavioral tasks. We use ten-fold cross-validation in the prediction analysis. Concretely, a random division of data folds is performed, and 90% of the data are used for training and 10% are used for testing. The predictive model built on the training set is directly tested on the testing set, and the Pearson correlation coefficient between actual and predicted scores is used to estimate the interpretability of learned representations for cognition. We repeat the ten-fold cross-validation procedure 20 times and report the averaged performance in Fig. 3. We make the following observations:

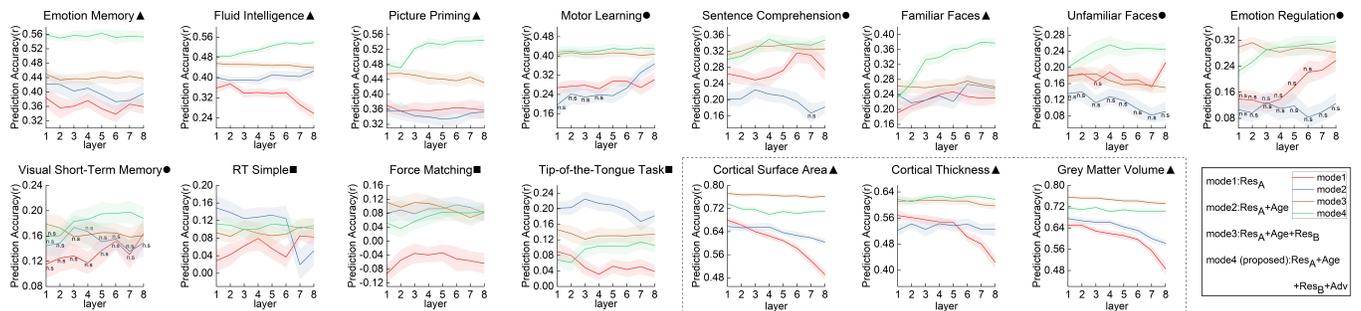

Fig. 3. Prediction results for representations and behavior scores/morphological estimates (inside the box). Note that ▲ denotes there are significant correlations for every mode in every layer, ■ denotes that all the correlations are not significant, and ● denotes that some of the correlations are not significant (labeled n.s).

All correlations between cognitive functions (involving comprehension, reasoning, memory, emotion, and learning) and representations extracted by MCIAT are statistically significant at false discovery rate (FDR)-corrected p < 0.05 (except for the VSTM task at the first layer; see details in Fig. 3), while the correlations of some basic active tasks (RT Simple, Force Matching, and TOT) are not significant at FDR-corrected p > 0.05. Specifically, Emotion Memory shows the highest predictability, with mean correlations between actual versus predicted scores reaching $r \geq 0.559 \pm 0.012$ for features in every transformer layer, averaging across all cross-validation repetitions. For other metrics scores of cognitive functions, the correlations vary from $r(VSTM) \geq 0.159 \pm 0.021$ to $r(FI) \geq 0.484 \pm 0.013$, and most of the correlations tend to increase as the transformer layers become deeper. The significant interpretability demonstrates that MCIAT encodes cognition-relevant features and represents intrinsic properties of the brain.

### 2) Association between Representation and Cognition When Controlling the Effect of Age

To verify that the proposed MCIAT learned age-independent cognitive information, we excluded age-association features in model building. Concretely, we first computed the Pearson correlation between individual encoded features and ages, and the features whose correlation survived the Bonferroni correction were removed (p < 0.05). Then, the individually remaining features were used to build prediction models, and repetitions of ten-fold cross-validation were performed, as described in Section IV-C.

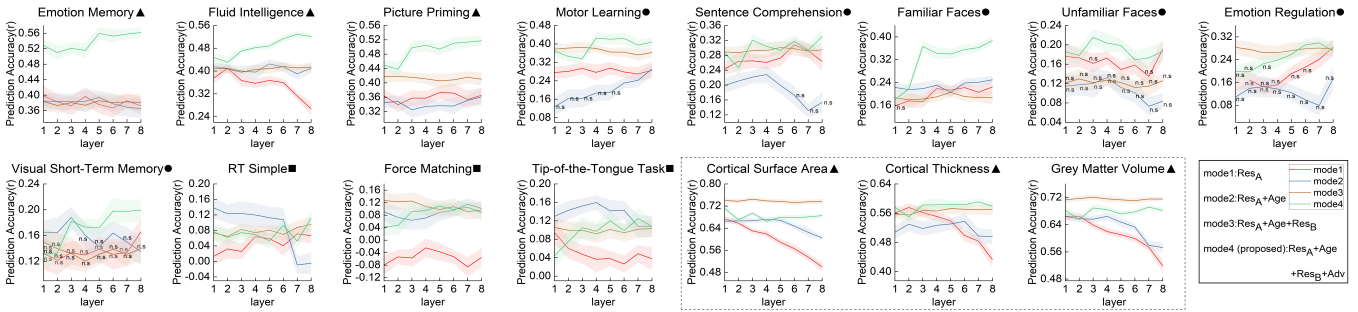

Fig. 4. Prediction results for representations and behavior scores/morphological estimates (inside the box) when controlling for the effect of age. Note that ▲, ■, and ● have the same meaning as in Fig. 3.

After controlling for the effect of age, correlations of almost all cognitive function scores were still significant at FDR-corrected p < 0.05 (except for the Emotion Regulation task at the first three layers and the VSTM task at the first two layers; see details in Fig. 4), and the correlations of basic active tasks were still not significant (FDR-corrected p > 0.05). This result suggests that our proposed MCIAT encodes cognition-relevant representations independent of age. Notably, for the first three layers of the model, the prediction performance slightly decreases after controlling for the effect of age, while there is little or no effect on the deeper layers, which indicates that the ViT encoder may tend to extract low-level age-related features in the first few layers and encode high-level cognition-relevant semantic information in the deeper layers.

### 3) Association between Representation and Morphological Estimates

We study the association between the brain representations and three common morphological estimates, including cortical surface area, cortical thickness and GM volume, using the abovementioned repetitive ten-fold cross-validation framework. The results with/without controlling the effect of age are illustrated in Fig. 3 and Fig. 4. There are significant correlations between the encoded features and morphological estimates (FDR-corrected $p < 2.0 \times 10^{-21}$), and the prediction performance trends oscillate downward as the transformer layers become deeper. In this context, the superior performance of MCIAT in downstream tasks can be theoretically interpreted from the following two perspectives: (1) The pre-trained encoder represents both morphological features and high-level semantic information of the brain, thus making a holistic comprehension of fine-grained structural characteristics and cognitive functions. (2) Each transformer block has different preferences for the representing of the brain, so selecting and fusing the important features of each layer in fine-tuning, which fosters the combination of multi-level, multi-stage representations, can facilitate the classification performance of the model.

### 4) External Validation on an Independent Dataset

Considering the broad replication crisis in neuroscience, especially in brain structure-behavior studies [51], external validation on an independent cohort is required to reliably identify the association between brain representations and cognition. We transfer the FI predictive model trained on the Cam-CAN dataset to the CNP dataset to predict the scores of two WAIS-IV subtests: letter number sequencing and matrix reasoning, which are the most relevant to fluid intelligence. (The standard form of the Cattell Culture Fair is used to measure the "fluid intelligence" for subjects from the Cam-CAN dataset, while the subjects from the CNP dataset are tested by WAIS-IV.) We observe a significant correlation between actual and predicted scores, as shown in Fig. 5. The transfer-validation result demonstrates that MCIAT has learned cognition-relevant representations and enables conceptual replications and extrapolations for subsequent research.

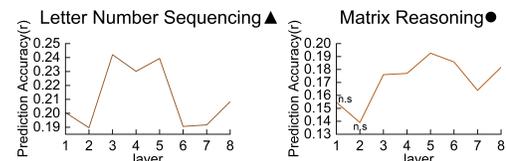

Fig. 5. Transfer results on CNP. ▲, ● and n.s are the same as in Fig. 3.

*5) Ablation Study for Representation Analysis*

We conduct a representation analysis ablation study to explore how each component affects the representation learning preferences of MCIAT. The results are illustrated in Fig. 3 and Fig. 4. We make the following observations: (1) Significant correlations with cognition were found for the encoded features extracted from the model pre-trained by a single proxy task, parallel distillation masked image modeling (ablation mode 1). This suggests that after incorporating a biologically meaningful prior, a simple mask-reconstruction task could enable the model to capture high-level semantic information from the brain structure. (2) The age prediction task weakens the correlations of representations with some cognitive functions but improves their correlation with morphological estimates, suggesting that age prediction, as an auxiliary proxy task, enhances the model's ability to learn structural information but makes the model prone to overfitting and may impair the model's understanding of overall semantics. (3) Visible portion restoration (restorative learning branch B) improves both the correlations of learned representations with cognitive functions and morphological estimates, suggesting that such a task can serve as a basic self-supervised pre-training component in a representation learning framework. Previous studies have indeed taken this transformation-restoration approach [9, 28, 52], although they have not made considerations based on comprehensive representation analysis. (4) The addition of adversarial learning significantly improves the correlations between representations and cognitive functions, suggesting that adversarial learning, especially the synergy between adversarial and restorative learning, facilitates the model to capture high-level semantics and learn general brain representations. Surprisingly, previous computer vision literature has often assumed that adversarial learning can enhance the ability of models to capture spatial features such as shape and texture [53, 54], but we reach a different conclusion. The fact that adversarial learning reduces the correlations between representations and morphological estimates of the brain exemplifies the heterogeneity of natural and medical images and highlights the importance of re-exploring the impact of various pre-training methods on model capabilities in brain representation learning.

TABLE V
LINEAR-PROB ACCURACY (%) ON ADHD-200

| Layer | 1 | 2 | 3 | 4 | 5 | 6 | 7 | 8 |
|---|---|---|---|---|---|---|---|---|
| mode1 | 60.23 | **60.82** | 57.31 | 57.89 | 59.65 | 58.48 | 56.14 | 59.06 |
| mode2 | 61.40 | 59.65 | **64.33** | 62.57 | 56.14 | 62.57 | 63.74 | 62.57 |
| mode3 | 60.82 | **63.16** | 62.57 | 56.14 | **63.16** | 57.31 | 59.06 | 60.82 |
| mode4 | 62.57 | 61.40 | 60.23 | 57.31 | 63.16 | **66.67** | 57.89 | 59.65 |

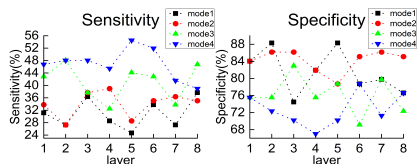

Fig. 6. Linear-prob sensitivity and specificity of ADHD-200.

*6) Linear-prob Classification on ADHD-200*

The linear-prob classification results of ADHD-200 are shown in Table V. In particular, we freeze the pre-trained models as the feature extractors and feed them with all the data in the ADHD-200 training and testing sets. Then, the encoded features of the training set are used to train a linear SVM as a classifier, which is tested on the encoded features of the testing set in each transformer layer. Our proposed MCIAT achieves a 66.67% linear-prob accuracy, surpassing most of the compared methods, indicating that the pre-trained encoder has learned how to represent biological patterns of the brain organization in healthy people and thus can discriminate intra-class variation and inter-class variation. Notably, we visualize the tendencies of sensitivity and specificity in Fig. 6, finding that the model pre-trained by a simple proxy task (ablation mode 1) tends to identify most of the test samples as healthy people (SPE > 74%, SEN < 38%), echoing our perspective that a simple optimization objective does not allow the model to extract sufficiently discriminative features. In fact, the sensitivity of MCIAT increases as the pre-training components are added one by one, suggesting that each component makes the model more sensitive to disease-related alterations, and the important role of multi-task collaborative pre-training was again emphasized.

### D. Representation Analysis for the Fine-tuned Models

In Section IV-C, we have observed that: (1) Correlations between MCIAT-extracted representations and cognitive functions tend to strengthen with increasing depth of the transformer layers, whereas for morphological estimates, the correlation strength tends to oscillate downward or is relatively stable, whether or not the effects of age are controlled for. (2) After controlling for the effect of age, correlations between representations and cognitive functions slightly decrease in the first three layers, while there is little or no effect in deeper layers. Therefore, we theoretically answer the question that features from different transformer layers represent different brain representations, so feature selection and fusion is more suitable for downstream tasks than using features from the last layer. Here, we further validate this observation in the fine-tuning stage and demonstrate the validity of the proposed MATS method from the perspective of representation analysis.

*1) Selected Frequency and Location of Tokens*

We computed the selected frequency of each token in each transformer layer (what percentage of subjects selected it) using the model that performed best in ADHD classification. Fig. 7 (a) and (b) show that the distribution of selected tokens is wide and discrete, and dominant tokens are very scarce, indicating that there is no strong consistency among individuals.

Moreover, Fig. 7 (c) shows that the selected tokens are different between NC and ADHD, and the differences increase layer by layer. This demonstrates that MCIAT can locate informative regions, even if they are displaced by disease or cognitive variation, and later separate patients from healthy people using the selected and fused discriminative features.

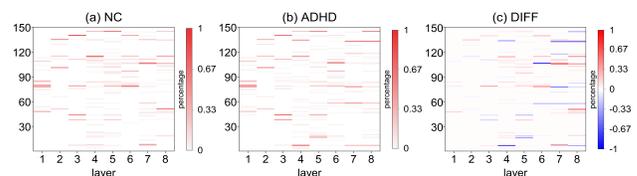

Fig. 7. Selected frequency of each token in each layer.

In Fig. 8, we choose 6 subjects (3 ADHDs, 3 NCs) and visualize the locations of the tokens that are the first to be selected for each subject in each layer. It suggests that there are both inter-class variation and intra-class variation, indicating that the token selection strategy is highly individualized.

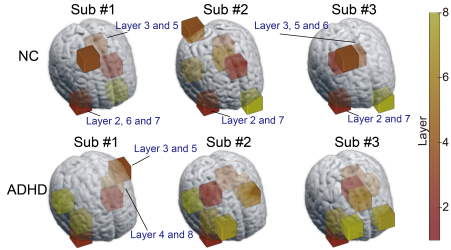

Fig. 8. The locations of the first-selected tokens in each transformer layer.

### 2) Visualization in the Latent Space

We visualize all 150 tokens and three MATS-selected tokens for each transformer layer in the latent space to demonstrate the validity of the MATS method. Specifically, we choose the models that performed best in ADHD classification for ablation mode 4 (w/o IAT) and ablation mode 4 (proposed) and visualized the distribution of ADHD and NC in the latent space via t-Distributed Stochastic Neighbor Embedding (t-SNE) for features in each layer. The results are illustrated in Fig. 9. We can observe that after adding IAT, which is implemented by our proposed MATS method, ADHDs and NCs are better separated in the latent space, especially for the last four layers. Notably, for ablation mode 4 (w/o IAT), the classifier is fed with all 150 tokens in the last layer, while ADHDs and NCs are not roughly separated even in the last layer, which explains the large gap in classification accuracy (69.01% vs. 74.27%) between ablation mode 4 (w/o IAT) and ablation mode 4 (proposed) to some extent. Therefore, we demonstrate that MATS not only fuses the features from different layers (different brain representations) but also reduces redundant features to prevent the model from being confused by disease-independent features, making it more suitable for downstream tasks.

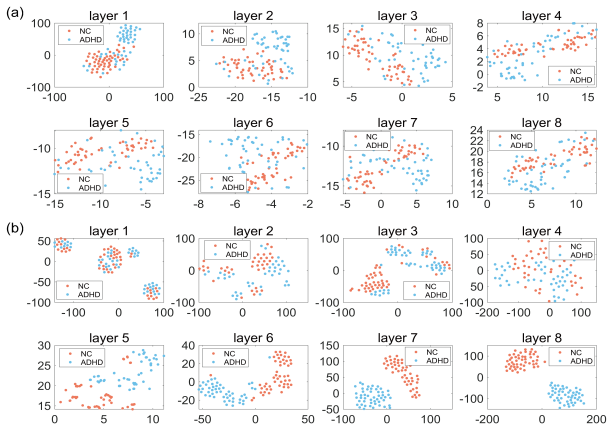

Fig. 9. t-SNE visualization. (a) Latent space of ablation mode 4 (w/o IAT). (b) Latent space of our proposed model.

## V. CONCLUSION

We propose MCIAT, a unified end-to-end framework for brain representation learning, combining collaborative pre-training and individual-adaptive-tokens fine-tuning. The key insight is that capturing fine-grained details and understanding how they relate to the overall global structure are of equal importance. The MCIAT shows remarkable performance compared to previous studies. We study 12 behavioral tasks and demonstrate that MCIAT-extracted representations have significant associations with cognitive functions and three morphological estimates, verifying the interpretability of MCIAT. Analysis of the selected frequency, location, and t-SNE visualization of tokens indicate that the model flexibly captures a set of discriminative regions, no matter how individually their location changes, reflecting the importance of a suitable strategy (i.e., IAT) for feature selection and fusion.

There are several inspirations to be considered in the future. First, we employ only one fixed patch size in this study, and it is reasonable to explore the influence of patch size to determine the optimal one. Second, we verify the effectiveness of the MCIAT on multiple classification tasks, while a wide range of downstream tasks (i.e., brain tumor segmentation, multiple MRI modality synthesis, etc.) may make the performance of the model more convincing. Moreover, given the interpretability of the learned representations, it is a natural idea to generate explainable biological guides by the pre-trained model to reinforce the fine-tuning procedure. We envisage that MCIAT will inspire future work on collaborative training, feature selection and feature fusion to improve the learning of universal representations for medical imaging.